\documentclass[aps,prx,onecolumn,longbibliography]{revtex4-1}

\usepackage{amssymb, amsmath, amsfonts, amsthm, mathtools}
\usepackage{setspace}
 
\usepackage{comment}
\usepackage{hyperref}
\usepackage{graphicx}
\usepackage{dcolumn}
\usepackage{xcolor}
\usepackage{multirow}
\graphicspath{ {images/} }

\usepackage{bm}

\begin{document}

\preprint{APS/123-QED}

\title{ Frequency-Based Reservoir Computing}

\author{Arthur S Powanwe}
\email{spowanwe@uwo.ca}
\affiliation{Department of Mathematics, Western University, London, ON, Canada}


\date{\today}

\begin{abstract}
Reservoir computing has emerged as an efficient machine learning framework for predicting time series generated by dynamical systems. In contrast to other machine and deep learning approaches, a reservoir computer trains only the output layer via linear regression, leaving the reservoir (recurrent layer) untrained. This simplification makes reservoir computers easier to train and more amenable to experimentation. However, because current reservoirs consist of networks of randomly connected nodes and require the optimization of numerous hyperparameters, a framework that precisely explains how reservoir computing operates and how it can be optimized remains missing. Here, we propose a frequency-based reservoir inspired by the brain's oscillatory dynamics and its hierarchy of timescales.
The frequency-based reservoir can be interpreted as an ensemble of independent oscillatory units, each processing a portion of the input's frequency content.
 This allows us to understand the reservoir's internal behavior by modeling it as a single unit driven by an external input. Borrowing from the theory of a nonlinear oscillator forced by complex periodic inputs, we found that units of the frequency-based reservoir selectively amplify and store specific input frequencies, which are then used for prediction. The frequency-based reservoir performs as well as or better than equivalent random reservoirs.
Furthermore, the frequency-based approach can be optimized to improve short-term prediction, a property that random reservoirs lack. Finally, we show that the frequency-based reservoir can also predict complex spatiotemporal dynamics. Our results show that reservoir computing can be designed using brain properties and theoretical insights borrowed from the physics of forced nonlinear oscillators.
\end{abstract}

\maketitle
\newpage

\section{Introduction}
Reservoir computers are a class of neural networks that consist of three main modules: an input layer that maps the input to the reservoir; a recurrent layer (the reservoir) that stores information about the input; and an output layer that maps the reservoir to a target \cite{jaeger2001echo,jaeger2004harnessing,jaeger2007optimization}. The input and recurrent layers are initialized randomly and kept fixed, whereas only the output layer is trained, typically using a simple linear regression. This provides a significant advantage for reservoir computers over other machine learning models that require training the recurrent layer. Despite their simple training, reservoir computers have been shown to perform surprisingly well at forecasting time series generated by chaotic dynamical systems \cite{lukovsevivcius2009reservoir,tanaka2019recent}. 
Understanding how the reservoir encodes information about the input to generate a prediction is crucial for designing optimal reservoir computers. Unfortunately, apart from initializing the recurrent layer at the edge of chaos,  there is currently no methodology for designing optimal reservoir computers. Building reservoir computers that clarify how the recurrent layer processes the input to generate a prediction will not only enhance interpretability but also enable effective implementation.\\

Recent works have focused on making reservoir computers more interpretable by incorporating brain-like properties. This includes building biologically motivated reservoirs that contain excitatory and inhibitory neurons \cite{maass2002real}, with connections between nodes subject to synaptic plasticity. It was demonstrated that controlling the excitatory-inhibitory (E-I) balance in a brain-inspired reservoir could significantly boost its performance \cite{srinivasan2025boosting}. Similarly, increasing the number of connections between neurons has been shown to improve accuracy. Recently, oscillations have been introduced as a critical substrate for computations and learning \cite{ceni2024random,effenberger2025functional,kramer2025brain}. Oscillations have been observed in the brains of several species and across diverse frequency bands \cite{buzsaki2006rhythms}. They usually exhibit a substantial variability, appearing as short, random epochs of synchrony \cite{xing2012stochastic,burns2011gamma}. They are believed to be involved in many cognitive processes \cite{wang2010neurophysiological}, including communication \cite{fries2005mechanism,fries2015rhythms,palmigiano2017flexible}, attention\cite{bosman2012attentional}, and working memory \cite{yamamoto2014successful,lundqvist2016gamma}. However, it remains poorly understood how neural oscillations contribute to computations and learning. Recent reservoir computing approaches that use oscillations have been shown to outperform traditional random reservoirs \cite{ceni2024random} and to be more interpretable \cite{kramer2025brain}. Nevertheless, those reservoirs still employ a random recurrent layer, which makes them difficult to optimize. Here, we introduce a reservoir computing approach based on the diversity of oscillatory behavior across frequency bands and brain regions. The reservoir comprises excitatory and inhibitory populations, each oscillating at a distinct frequency. It uses the frequency property of units for computation; hence the name "frequency-based reservoir computer". In addition, the frequency-based reservoir computer is endowed with the hierarchy of timescales \cite{hasson2008hierarchy,murray2014hierarchy,runyan2017distinct,spitmaan2020multiple,siegle2021survey,manea2022intrinsic,li2022hierarchical}, an important and widespread property of brain computation.\\

The frequency-based reservoir computing approach is motivated by the oscillatory dynamics of the Wilson-Cowan model \cite{wilson1972excitatory,wallace2011emergent}. Each unit consists of two nodes, each representing a population of excitatory or inhibitory neurons. The model was designed to ensure that each unit oscillates at a specific frequency, behaves as a brain region with its own timescale, and exhibits the noisy behavior inherent in brain recordings. In contrast to previous works \cite{ceni2024random,effenberger2025functional,kramer2025brain}, the coupling between the units is not chosen randomly. Furthermore, we found that coupling between the units was optional, suggesting that computations happen at the unit level. With this choice, the frequency-based reservoir can be viewed as an ensemble of small, independent reservoirs (each comprising two nodes) that process a portion of the input's frequency content. Borrowing from the theory of a nonlinear oscillator forced by a complex input, we found that the reservoir units selectively amplify frequency components of the input that are near their intrinsic frequencies. A property that is lacking in random reservoirs. We further found that frequency-based reservoirs perform as well as, or better than, equivalent random reservoirs and can be optimized to improve short-term prediction performance. Finally, we extend the predictive capabilities of frequency-based reservoirs to complex spatiotemporal chaotic inputs. Taken together, our results show that brain-inspired reservoir computing can be designed to be fully interpretable and optimized for a specific task.

\section{Results}
\subsection*{Frequency-based reservoir computing model}
We consider the dynamics of the reservoir computer obeying the following equation

\begin{equation}
\label{reservoir_dynamics}
\text{r(t+1)}=(1-\alpha)\text{r(t)}+\alpha \tanh \big(\text{A} \text{r(t)}+\text{W}_\text{in}\text{u(t)}\big).
\end{equation}

\begin{figure*}
\includegraphics[height=3.0in,width=6.5in]{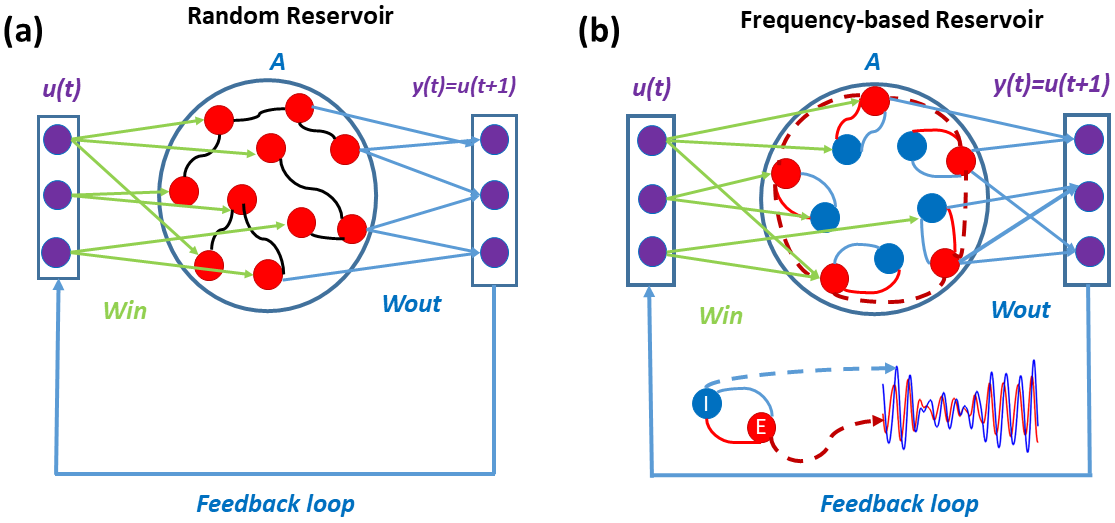}
\caption{\textbf{Two reservoir computing approaches with the same architecture but distinct recurrent layers.} $\mathbf{(a)}$ In a random reservoir, the recurrent layer is initialized by randomly choosing couplings among the nodes. $\mathbf{(b)} $ In a frequency-based reservoir, the recurrent layer consists of coupled excitatory and inhibitory nodes that can be coupled or not. Each unit has intrinsic oscillatory dynamics with a specific frequency used for computation. For both reservoir types, the matrix $\mathbf{W_{in}}$  that maps the input to the recurrent layer is chosen randomly, and the output matrix $\mathbf{W_{out}}$ that maps the recurrent layer dynamics to the target is obtained through linear regression.}
\label{FB_diagram}
\end{figure*}

The matrix $\text{A}$ describes the connectivity among the nodes and is usually chosen to be a random matrix (Fig.\ref{FB_diagram}-(a)). The matrix $ \text{W}_\text{in}$ maps the lower-dimensional input $\text{u(t)}$ to the higher-dimensional recurrent layer and is also a random matrix. The coefficient $\alpha$ is a leaky rate that controls how quickly the variable $\text{r(t)}$ incorporates the reservoir's memory of past values. The goal in reservoir computing is to train an output layer $\text{W}_\text{out}$ so that the dynamics of the reservoir follow a target signal $\text{y(t)}$. In this work, we are interested in the prediction of the next step  $\text{y(t)=u(t+1)}$ of the input $\text{u(t)}$. During training, we minimize the following quantity.

\begin{equation}
\label{loss_function}
   \| \mathbf{W_{out}r(t)-y(t)} \|_2
\end{equation}

This is done using a linear regression. During testing, we use the target as input, that is we add a feedback loop (Fig.\ref{FB_diagram}-(a)) so that the input is now $\text{W}_\text{out}\text{r(t)}$ and the reservoir becomes a self-evolving dynamical system that projects to the target via $\text{W}_\text{out}$. Random reservoirs (Fig.\ref{FB_diagram}-$\mathbf{(a)}$) have been shown to perform very well in chaotic timeseries forecasting \cite{jaeger2004harnessing}. However, the reservoirs are often regarded as a black boxes, lacking interpretability and being difficult to optimize. The magic relies on hyperparameter optimization, a process that is also poorly understood and requires substantial computational resources rather than a transparent methodology. Here, we propose a reservoir computing framework that does not rely on random connections among nodes, but rather on the intrinsic oscillatory dynamics of each unit (Fig.\ref{FB_diagram}). Our reservoir framework is motivated by the Wilson-Cowan model, a well-known model for representing the dynamics of coupled populations of excitatory and inhibitory neurons in the brain \cite{wilson1972excitatory,wallace2011emergent}. A unit describes a system of two connected nodes, one node representing a population of excitatory neurons (red disc in Fig.\ref{FB_diagram}-$\mathbf{(b)}$) and the other a population of inhibitory neurons (blue disc in Fig.\ref{FB_diagram}-$\mathbf{(b)}$). We are interested in the oscillatory behavior of each unit in isolation and when coupled to other units. In the original Wilson-Cowan model (Supplementary Information), the E and I populations have a recurrent coefficient that captures interactions within each population. They receive constant external input, and coupling between units occurs only from excitatory populations. Two distinct types of coupling were chosen, a ring-graph coupling where each unit is only coupled to its nearest neighbor (see dashed red line Fig.\ref{FB_diagram}-$\mathbf{(b)}$ here dashed means optional) and a distance-dependent graph coupling where the strength of the coupling between units decreases as the distance between them increases. However, we found that the coupling mainly shaped each unit's intrinsic dynamics. Inspired by the original Wilson-Cowan dynamics, we proposed a version that follows the common reservoir-computing dynamics (Equation \ref{reservoir_dynamics}). Because computation happens at the unit level (see Supplementary Information), the coupling between the units is optional (dashed red line Fig.\ref{FB_diagram}-$\mathbf{(b)}$). We do not claim that coupling does not matter; it mainly shapes the intrinsic behavior of each local unit. In addition, for each coupled network, we can find a corresponding uncoupled one with similar properties.

\begin{figure*}
\includegraphics[height=3.0in,width=3.5in]{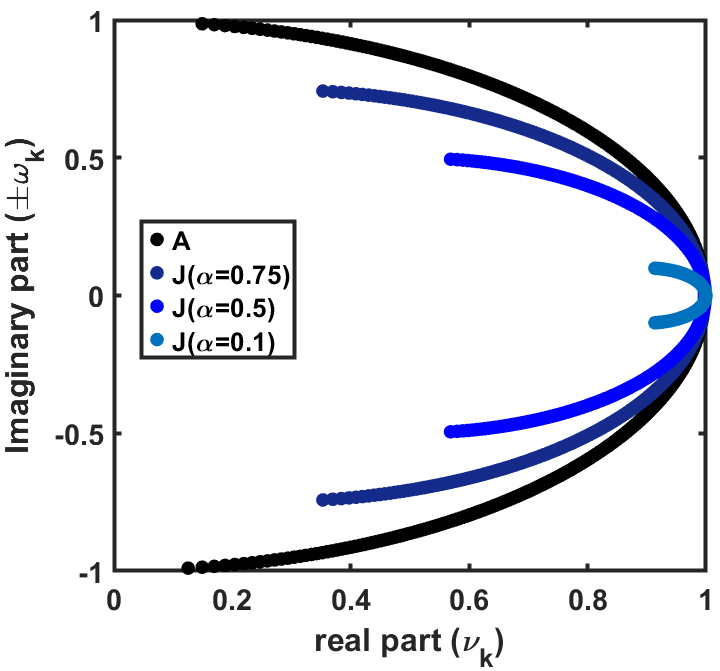}
\caption{\textbf{Spectra of the recurrent $\mathbf{A}$ and Jacobian $\mathbf{J}$ matrices for different values of the coefficient $\alpha$} The y axis shows the imaginary part and the x axis the real part of the eigenvalues. }
\label{FB_eigenspectrum}
\end{figure*}
In the frequency-based reservoir, the recurrent matrix $\mathbf{A}$ is chosen so that its eigenvalues are complex conjugate pairs (Fig.\ref{FB_eigenspectrum}.
Similarly, the eigenvalues of the  Jacobian matrix $\text{J}=(1-\alpha)\text{I}+\alpha\text{A}$ of Equation.\ref{reservoir_dynamics} are also complex conjugate pairs (Fig.\ref{FB_eigenspectrum}).
The recurrent matrix is block-diagonal, with each block representing the intrinsic dynamics of a unit.

\begin{equation}
\mathbf{A} = \left[\begin{array}{cccccccccc}
\nu_1 & \omega_{1} & 0 & 0 & \cdots & 0 & 0 & \cdots & 0 & 0 \\
-\omega_{1} & \nu_{1} & 0 & 0 & \cdots & 0 & 0 & \cdots & 0 & 0 \\
0 & 0 & \nu_{2} & \omega_{2} & \cdots & 0 & 0 & \cdots & 0 & 0 \\
0 & 0 & -\omega_{2} & \nu_{2} & \cdots & 0 & 0 & \cdots & 0 & 0 \\
\vdots & \vdots & \vdots & \vdots & \ddots & \vdots & \vdots & \cdots & \vdots & \vdots \\
0 & 0 & 0 & 0 & \cdots & \nu_{k} & \omega_{k} & \cdots & 0 & 0 \\
0 & 0 & 0 & 0 & \cdots & -\omega_{k} & \nu_{k} & \cdots & 0 & 0 \\
\vdots & \vdots & \vdots & \vdots & \cdots & \vdots & \vdots & \ddots & \vdots & \vdots \\
0 & 0 & 0 & 0 & \cdots & 0 & 0 & \cdots & \nu_{N} & \omega_{N} \\
0 & 0 & 0 & 0 & \cdots & 0 & 0 & \cdots & -\omega_{N} & \nu_{N}
\end{array}\right]
\end{equation}

\begin{figure*}
\includegraphics[height=4.0in,width=6.5in]{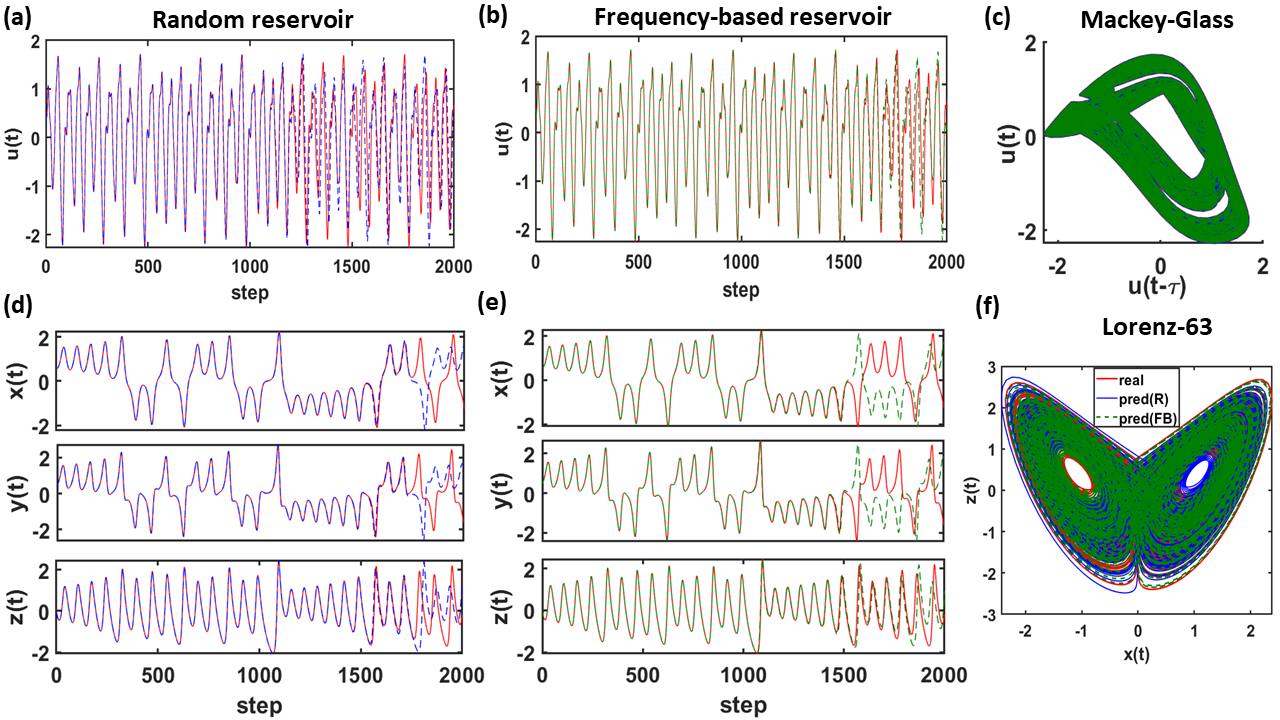} 
\caption{\textbf{Closed-loop forecasting of the Mackey-Glass and the Lorenz-63 timeseries and the corresponding attractors.} $\mathbf{(a)}$ Short-term prediction of the Mackey-Glass timeserie using the random and $\mathbf{(b)}$ the frequency-based reservoirs. $\mathbf{(c)} $ Long-term prediction of the Mackey-Glass attractor for both the random and frequency-based reservoirs.$\mathbf{(d)}$ Short-term prediction of the Lorenz-63 timeserie from the random and $\mathbf{(e)}$ Frequency-based reservoirs. $\mathbf{(f)}$ Long-term prediction of the Lorenz-63 attractor for both the random and frequency-based reservoirs. For all panels, the red data refer to the real data, while the blue and green data refer to the predictions from the random and frequency-based reservoirs, respectively.}
\label{FB_forecasting}
\end{figure*}

The coupling between the units is optional; therefore, we did not include it here. However, in the original Wilson-Cowan formulation, we considered distinct coupling matrices. Our recurrent matrix is not randomly generated; it can be selected appropriately based on the input we aim to forecast.
 We arranged the units so that the first unit has the lowest angular frequency, $\omega_{min}$, and the last unit, $\text{N}$, has the largest angular frequency, $\omega_{max}$.
 With this choice, each unit is a small reservoir of two connected nodes.  The frequency-based reservoir is an ensemble of $\text{N}$ small reservoirs (each having two nodes). Furthermore, the reservoir is endowed with a hierarchy of timescales \cite{hasson2008hierarchy,murray2014hierarchy,runyan2017distinct,spitmaan2020multiple,siegle2021survey,manea2022intrinsic,li2022hierarchical} (units with large indices $k$ have larger angular frequencies $\omega_\text{k}$ and appear early in the hierarchy. In contrast, those with small $k$ have smaller angular frequencies $\omega_\text{k}$ and appear later), a key property of brain computation. Finally, we added white Gaussian noise to the dynamics to model the stochasticity of neural recordings (see Methods).
To investigate the ability of the frequency-based reservoir computing to predict chaotic time series, we chose two dynamical systems known to generate chaotic dynamics, namely the Mackey-Glass \cite{mackey1977oscillation} and the Lorenz-63 chaotic model \cite{lorenz2017deterministic}. We evaluated random and frequency-based reservoirs under these two inputs and trained them to predict their next steps. This enables us to obtain the output matrix used during closed-loop testing to forecast future values. To maintain a fair comparison between the random and frequency-based reservoirs, we used the same Hyperparameters for both. The results are shown in Fig.\ref{FB_forecasting}, which displays only the data from the closed-loop forecasting phase.

We found that the frequency-based reservoir performed as well as, or better than, the random reservoirs under the same conditions. For the example in Fig.\ref{FB_forecasting}, we did not aim to optimize both reservoirs, and the hyperparameters were selected manually. The goal was to compare random and frequency-based reservoirs under identical conditions.  We consider a network size of \text{N=301} units (601 nodes) for the frequency-based reservoir. Note that for the first unit with an imaginary part equal to zero, we only chose one node instead of two to form a unit. The values of the component of the recurrent matrix $\text{A}$  are shown in Fig.\ref{FB_eigenspectrum}, and the coefficient $\alpha$ was set to be 0.5. The frequency-based reservoir is a promising candidate for time-series forecasting. By design, it has a very low number of components. For a matrix of size \text{2Nx2N}, the number of non-zero components is \text{4N}, and the number of distinct non-zero components is \text{2N}. In addition, each unit is independent of the others, and computation occurs at the unit level. This stands in clear contrast to a random reservoir, in which computations arise from the collective behavior of the nodes in the recurrent layer. The frequency-based reservoir, therefore, offers a unique opportunity to understand how the reservoir operates and which computations occur within the recurrent layer. We can clearly understand the computations inside the recurrent layer by studying the dynamics of a single unit driven by the input.

\subsection*{Selective-frequency amplification of multi-frequency periodic inputs as a mechanism for information storage}
\begin{figure*}
\includegraphics[height=4.0in,width=6.5in]{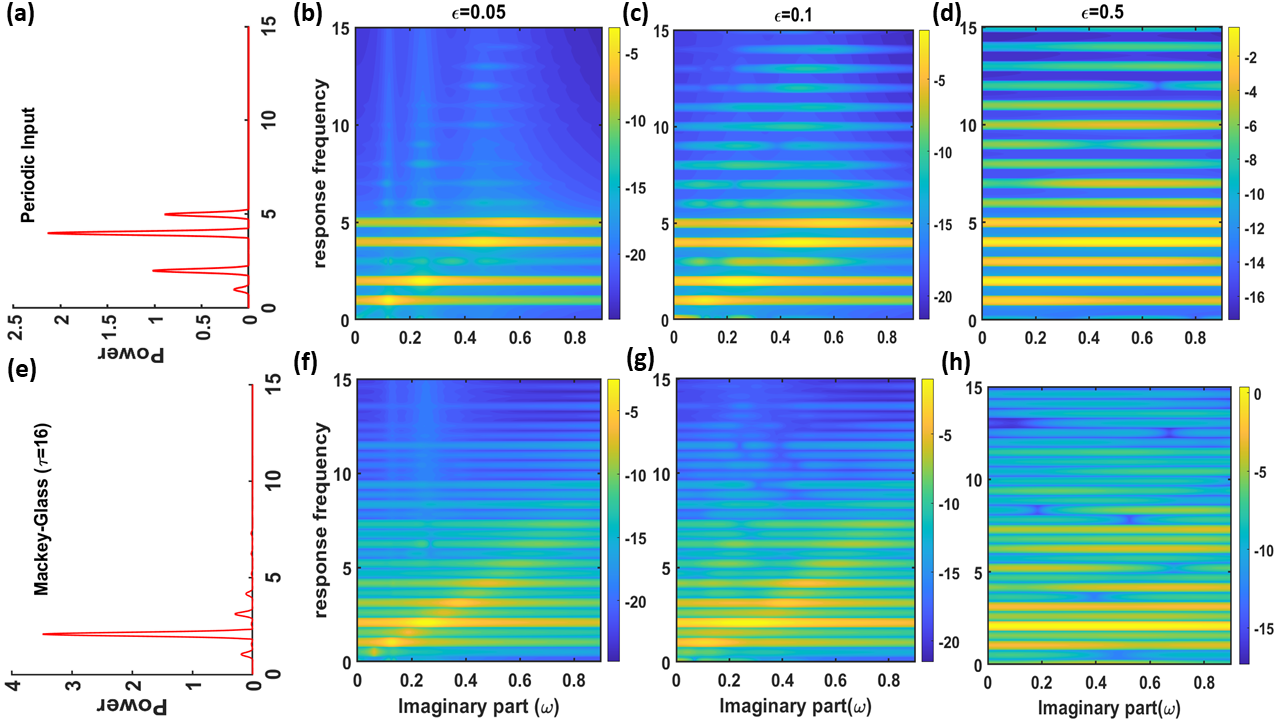}
\caption{\textbf{Selective frequency amplification of multi-frequency periodic inputs.} \textbf{(a)} Power spectrum of the multi-frequency periodic input described in Equation.\ref{periodic_input}. \textbf{(b)} \text{log(Power)} of a single unit, precisely Equation.\ref{reservoir_single_unit_node1} for low amplitude values $\epsilon=0.05$. \textbf{(b)}  \text{log(Power)} for intermediate amplitude values $\epsilon=0.1$. \textbf{(c)} \text{log(Power)} for large amplitude values $\epsilon=0.1$. \textbf{(e)} Power spectrum of the periodic Mackey-Glass  $(\tau=16)$ dynamical system. \textbf{(f)} \text{log(Power)} of a single unit with the Mackey-Glass as input for low amplitude values $\epsilon=0.05$.\textbf{(g)} \text{log(Power)}  for intermediate amplitude values $\epsilon=0.5$. \textbf{(h)} \text{log(Power)} for large amplitude values $\epsilon=0.5$. For all the panels concerned \textbf{(b)},\textbf{(c)},\textbf{(d)},\textbf{(f)},\textbf{(g)} and \textbf{(h)} the coefficients $\nu$ and $\omega$ were varied as in Fig.\ref{FB_eigenspectrum}. }
\label{Single_node_ampl}
\end{figure*}

To understand how the Frequency-based reservoir encodes and stores information about the input, let's study the dynamics of a single unit driven by a one-dimensional input $\text{u(t)}$. We write the dynamics of the Frequency-based reservoir for a single unit as follows
\begin{align}
     \label{reservoir_single_unit_node1}
    \text{r}_1\text{(t+1)}&=(1-\alpha)\text{r}_1\text{(t)}+\alpha\tanh\big(\nu \text{r}_1(t)+\omega \text{r}_2(t)+\epsilon_1 \text{u(t)}\big)\\
    \label{reservoir_single_unit_node2}
    \text{r}_2\text{(t+1)}&=(1-\alpha)\text{r}_2\text{(t)}+\alpha\tanh\big(-\omega \text{r}_1\text{(t)}+\nu \text{r}_2 \text{(t)}+\epsilon_2 \text{u(t)}\big)
\end{align}
To have a better understanding of how a single unit processes an arbitrary input $\text{u(t)}$, we first consider a periodic input with multiple frequencies defined as 
\begin{equation}
\label{periodic_input}
\text{u(t)}= \sum_{l=1}^{L} \text{A}_\text{l}\sin(\omega_\text{l }\text{t})
\end{equation}
The single unit of the frequency-based reservoir is a nonlinear oscillator that oscillates at a frequency given by the imaginary part of the corresponding Jacobian. The Equation. \ref{reservoir_single_unit_node1}-\ref{reservoir_single_unit_node2} and Equation.\ref{periodic_input} represent the dynamics of a nonlinear oscillator forced by a complex periodic input. Understanding the computation at a single unit is a physics problem borrowed from the theory of forced nonlinear oscillators. We investigate this question by analyzing the power spectrum of the system of Equation.\ref{reservoir_single_unit_node1}-\ref{reservoir_single_unit_node2} as the imaginary part of $\text{A}$ varies according to Fig.\ref{FB_eigenspectrum}. We are interested in the $log(Power)$  as a function of the frequency and the imaginary part $\omega$. We first consider a periodic input described in Equation. \ref{periodic_input}. The input is made of $\text{L=5}$  fundamentals frequencies $\text{f}_\text{l}=1,2,3,4, 5 \text{Hz}$ so that $\omega_\text{l}=2\pi \text{f}_\text{l}$. The amplitude  of each fundamental frequency is given by $\text{A}_\text{l}$ and is all scaled by a factor $\epsilon$. The corresponding Power spectrum is given in Fig.\ref{Single_node_ampl}-$\mathbf{(a)}$ and shows the contribution of each frequency to the input $\text{u(t)}$. The 4 \text{Hz} frequency component contributes the most, followed by the 5 and 2 \text{Hz} frequencies, then the 1 \text{Hz} frequency, and finally, the 3 \text{Hz} frequency is very weak due to the low value of $\text{A}_3$.
For low amplitude values $\epsilon=0.05$, a single unit responds to all the strong frequency components in the input independently of the value of its own intrinsic frequency (related to the imaginary part $\omega$). However, a unit powerfully amplifies the input's frequency components near its inherent frequency. The mechanism is similar to nonlinear resonance but involves multiple frequency components. Therefore, a unit oscillating at a given frequency will store more information about frequency components of the input that are nearby. We refer to this mechanism as selective frequency amplification. For intermediate amplitude values $\mathbf{\epsilon}=0.1$, the selective frequency amplification persists. Still, the unit receives more information overall about all frequency components, including the element with the lowest contribution, $\text{A}_3$. For a higher amplitude value $\mathbf{\epsilon}=0.5$, the selective amplification mechanism is destroyed, and the intrinsic oscillator frequency being near a frequency component of the input does not lead to additional amplification. The unit responds preferentially to the element with the highest amplitude. Therefore, components with higher amplitude are more amplified. In addition, we observe the emergence of higher-order harmonics due to nonlinear effects, resulting in increased amplitudes.
We consider a second periodic signal generated by the Mackey-Glass dynamical system with $\tau = 16$. The corresponding power spectrum Fig.\ref{Single_node_ampl}-$\mathbf{(e)}$ showed that the periodic Mackey-Glass can be seen as a periodic input with multiple fundamental frequencies of distinct amplitude, as in the first case Equation.\ref{periodic_input}. By also scaling the Mackey-Glass by a factor $\epsilon$, the analogy with Equation.\ref{periodic_input} is complete. For low amplitude values $\mathbf{\epsilon}=0.05$, the selective amplification is observed as previously described for the first periodic input. For intermediate amplitude values Fig.\ref{Single_node_ampl}-$\mathbf{(g)}$ selective frequency amplification persists and for high amplitude values Fig.\ref{Single_node_ampl}-$\mathbf{(h)}$ selective frequency amplification is destroyed. Selective amplification occurs for periodic inputs with low to intermediate amplitudes and is destroyed when the amplitude becomes large. Can selective frequency amplification be the mechanism at work in the frequency-based reservoir? To answer this question, we need to clarify a few points: 1) The mechanism we have described is for a periodic system that shows a Power spectrum with a discrete and finite number of frequencies. Chaotic systems, on the other hand, exhibit a continuous spectrum in which an infinite number of frequencies contribute to the dynamics. However, even in that case, some chaotic systems still exhibit frequencies that dominate their dynamics. It is then possible that the same mechanism generalizes for chaotic systems.

\begin{figure*}
\includegraphics[height=4.0in,width=6.5in]{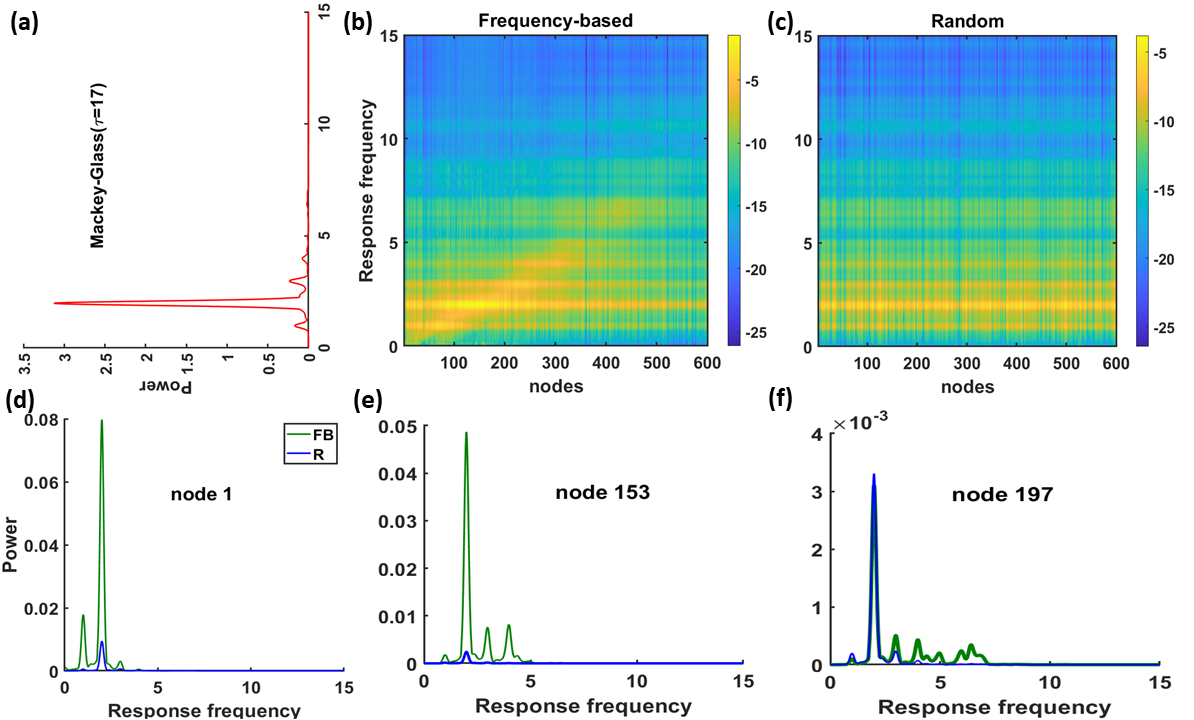}
\caption{\textbf{Selective frequency amplification of the chaotic Mackey-Glass input.} \textbf{(a)} Power spectrum of the chaotic Mackey-Glass $(\tau=17)$ dynamical system. \textbf{(b)} \text{log(Power)} of the frequency-based reservoir. \textbf{(c)} \text{log(Power)} of the random reservoir.\textbf{(d)} Power spectra of the first node in the frequency-based (green) and random (blue) reservoirs. \textbf{(e)} Power spectra of node 153 in the frequency-based (green) and random (blue) reservoirs. \textbf{(f)} Power spectra of node 197 in the frequency-based (green) and random (blue) reservoirs. }
\label{FB_ampl_MG}
\end{figure*}

\subsection*{Selective frequency amplification generalizes to a one-dimensional chaotic system}
Our goal is to understand how the frequency-based and random reservoirs process external inputs.
We consider the chaotic Mackey-Glass input used in Fig.\ref{FB_forecasting}-\textbf{(a-c)}. 
The delay in the Mackey-Glass equation introduces additional dimensions, rendering it an effective infinite-dimensional system. However, we refer to the dimensions as the number of variables simulated to generate the input. By this definition, the Mackey-Glass system is one-dimensional. In contrast to the periodic case $(\tau=16)$, the power spectrum of the chaotic case $(\tau=17)$ is continuous, meaning that an infinite number of frequencies contribute to the dynamics of the Mackey-Glass. Nevertheless, a finite number of frequency components contribute the most as shown in Fig.\ref{FB_ampl_MG}-\textbf{(a)}.
 We first consider the frequency-based reservoir. Each unit is uncoupled. Precisely, the dynamics of each unit in the recurrent layer are given by Equation.\ref{reservoir_single_unit_node1}-\ref{reservoir_single_unit_node2} with $\nu$,$\omega$ replaced by $\nu_k$ and $\omega_k$. In addition, the inputs amplitude $\epsilon_{1,2}$ in Equation.\ref{reservoir_single_unit_node1}-\ref{reservoir_single_unit_node2} are replaced by the corresponding coefficients $\text{W}_\text{in}(k)$. A total of 301 (N=601 nodes) of these units are considered with their parameters $\nu_k$ and $\omega_k$  covering all the spectrum shown in Fig.\ref{FB_eigenspectrum}.    
 The nodes in the reservoir were arranged so that the first node corresponds to the lowest intrinsic frequency (lowest value of $\omega_k$) and the last node to the largest intrinsic frequency (most significant value of $\omega_k$). This arrangement facilitated interpretation and aligned with the single-unit case.
 We compute the power spectrum of each node, and we display the \text{log(power)} as a heat map in Fig.\ref{FB_ampl_MG}-$\mathbf{(b)}$. We found that nodes in the reservoir selectively amplify frequencies near their intrinsic frequencies, a mechanism we have identified as selective frequency amplification in the context of multifrequency periodic inputs. For the chaotic input here, the selective frequency amplification generalizes to the continuous power spectrum of the Mackey-Glass input (Fig.\ref{FB_ampl_MG}-$\mathbf{(b)}$).   We further investigate the power spectra of individual nodes. The first node has the lowest intrinsic frequency and amplifies the dominant frequency component, the lowest component of the input, while the fastest components are dampened (Fig.\ref{FB_ampl_MG}-$\mathbf{(d)}$). Note that the Mackey-Glass equation has a strong frequency component that is always amplified independently of the node. Node 153 has an intermediate intrinsic frequency and amplifies intermediate- to high-frequency components of the input, but not low-frequency components. Finally, node 197 exhibits a high intrinsic frequency and selectively amplifies high-frequency components of the input while attenuating lower-frequency components.  For the random reservoir (Fig.\ref{FB_ampl_MG}-$\mathbf{(c)}$), there is no identifiable mechanism. All nodes process the input similarly. In contrast to the frequency-based reservoir, all the nodes preferentially process the largest frequency components present in the input. This is clearly evident in the power spectra of nodes 1, 153, and 197, which all exhibit a strong value at the same frequency.

\begin{figure*}
\includegraphics[height=4.0in,width=6.5in]{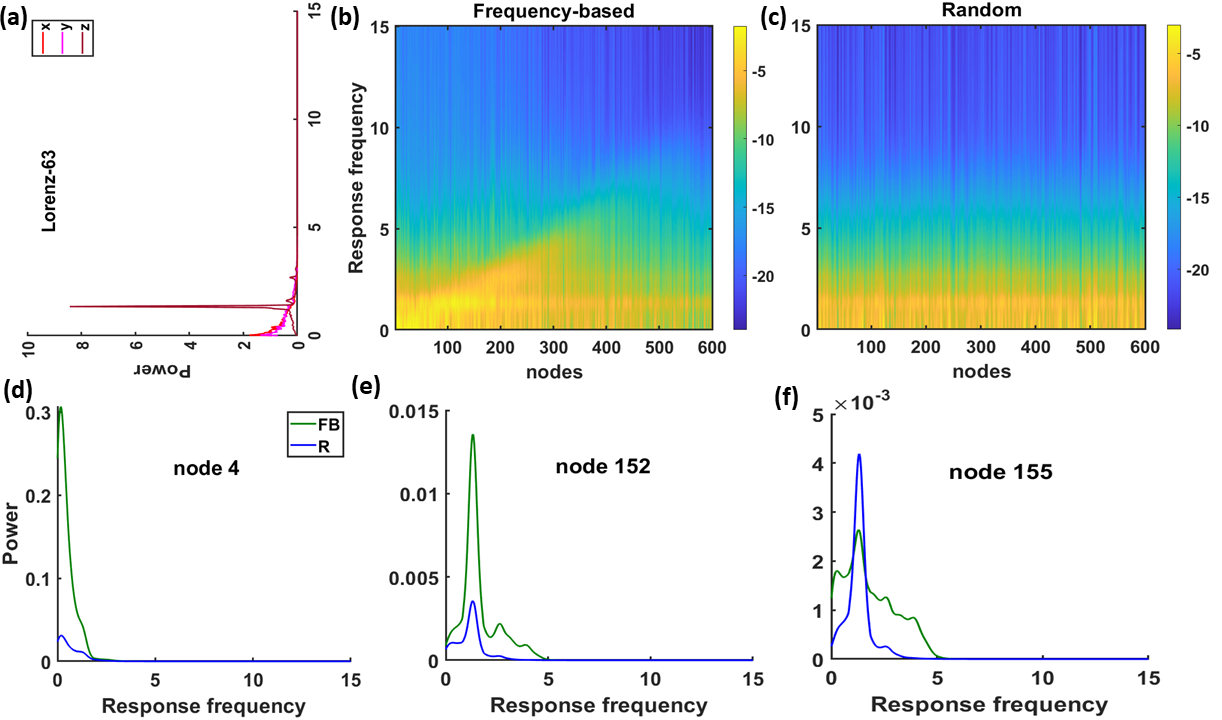}
\caption{\textbf{Selective frequency amplification of a multi-dimensional chaotic system} \textbf{(a)} Power spectra of the x,y and z components of the chaotic Lorenz-63 dynamical system.\textbf{(b)} \text{log(Power)} heat map of the frequency-based reservoir.\textbf{(c)} log(Power) heat map of the random reservoir. \textbf{(d)} Power spectra of node 4 in the frequency-based (green) and the random (blue) reservoirs. \textbf{(e)} Power spectra of node 152 in the frequency-based (green) and random (blue) reservoirs. \textbf{(f)} Power spectra of node 155 in the frequency-based (green) and the random (blue) reservoirs.}
\label{FB_ampl_Lorenz63}
\end{figure*}

\subsection*{Selective frequency amplification generalizes to a multi-dimensional chaotic system}
We now focus on the case of the three-dimensional Lorenz-63 chaotic system 
  used as the input to the frequency-based and random recurrent layers of the reservoir computers ( Fig.\ref{FB_forecasting}-\textbf{(d,e,f)}). The power spectra of the \text{x,y} variables are similar and concentrated at lower frequencies. They are broadband, with no clear dominant component. In contrast, the z component shows a strong peak at intermediate frequency values, characteristic of a dominant periodicity (Fig.\ref{FB_ampl_Lorenz63}-\textbf{(a)}). Overall, the Lorenz-63  dynamical system exhibits low, intermediate (dominant), and large frequencies. We analyze how the frequency-based and random reservoirs process the Lorenz-63 dynamical system.
  As in the one-dimensional input case, each unit in the frequency-based reservoir is decoupled from other units. Their dynamics are described by the Equation.\ref{reservoir_single_unit_node1}-\ref{reservoir_single_unit_node2} with $\nu$,$\omega$ replaced by $\nu_k$, $\omega_k$ covering the spectrum described in Fig.\ref{FB_eigenspectrum}. We found that the inputs to each odd and even component of a unit were $\text{W}^1_\text{in}(1,:,k)\text{x}+\text{W}^1_\text{in}(:,2,:,k)\text{y}+\text{W}^1_\text{in}(:,3,k)\text{z}$ and  $\text{W}^2_\text{in}(1,:,k)\text{x}+\text{W}^2_\text{in}(:,2,:,k)\text{y}+\text{W}^2_\text{in}(:,3,k)\text{z}$ respectively. Moreover, the x and y components are similar, and the effective input contribution were approximated to $(\text{W}^1_\text{in}(1,:,k)+\text{W}^1_\text{in}(:,2,:,k))\text{(x+y)}\text{W}^1_\text{in}(:,3,k)\text{z}$ and  $(\text{W}^2_\text{in}(1,:,k)+\text{W}^2_\text{in}(:,2,:,k))\text{(x+y)}+\text{W}^2_\text{in}(:,3,k)\text{z}$ respectively. This allows us to determine the amount of input from the x-, y-, and z-components that each unit receives. The situation, nevertheless, is similar to the multi-frequency input in Equation \ref{periodic_input}. The mostly lower frequency components of the x and y variables will be scaled by the amplitude  $(\text{W}^1_\text{in}(1,:,k)+\text{W}^1_\text{in}(:,2,:,k))$ and  $(\text{W}^2_\text{in}(1,:,k)+\text{W}^2_\text{in}(:,2,:,k))$ while the frequency components of the z variable will be scaled by $\text{W}^1_\text{in}(:,3,k)$ and $\text{W}^2_\text{in}(:,3,k)$. We found that the units in the reservoir selectively amplified frequency components close to their intrinsic frequencies while dampening other frequencies (depending on their amplitudes).
  The frequency-based reservoir employs the selective frequency-amplification mechanism (Fig.\ref{FB_ampl_Lorenz63}-\textbf{(c)}). Precisely, the node 4 in the Frequency-based reservoir selectively amplifies lower frequency components ( mostly x and y variables) of the Lorenz-63 input (see Fig.\ref{FB_ampl_Lorenz63}-\textbf{(d)}) and dampens higher ones. This can be explained by the fact that the frequency of the fourth node is low, the amplitudes to the x and y variables are strong enough, and the amplitude to the z variable is weak. Nodes (152 and 155) with intermediate or higher intrinsic frequencies selectively amplify the strong frequency component of the z variable and the higher frequency components of the three variables. Therefore, the selective frequency amplification mechanism generalizes to multidimensional chaotic systems. In the random recurrent layer, we found no discernible structure or transparent mechanism for processing input frequency. As in the one-dimensional case, all nodes responded similarly to the three-dimensional input.

\subsection*{Selective frequency amplification as a mechanism for chaotic time series prediction}
The frequency-based reservoir processes temporal information differently from the random reservoir via a selective frequency-amplification mechanism. The question becomes how the reservoir takes advantage of this mechanism to make an accurate prediction of chaotic dynamics, as we showed in Fig.\ref{FB_forecasting}.
In the reservoir computing framework, only the readout layer, $\text{W}_\text{out}$, is trained using linear regression. The information stored in the reservoir is crucial for accurate predictions. Previous attempts to explain how the reservoir stores information about the input have failed because of its random nature.  The frequency-based reservoir does not have this curse. Each unit in the reservoir stores a portion of the input information in the frequency domain. Specifically, units with lower intrinsic frequencies will store lower-frequency components of the input, units with intermediate intrinsic frequencies will store intermediate-frequency components, and units with higher intrinsic frequencies will store higher-frequency components. In addition, strong frequency components of the input are stored in all the units. The frequency-based recurrent layer acts as a set of many small reservoirs. Each small reservoir is a unit (a layer of two nodes) that stores part of the input information. This is similar to an ensemble of small reservoirs, each processing a part of the input in the frequency domain. When the reservoir is sufficiently large and the stimulation duration is sufficiently long, all information about the input is stored in the recurrent-layer units. A simple linear combination $\text{W}_\text{out}\text{r(t)}$ of all the units should be enough to recover all the information about the input and therefore make an accurate prediction. In addition, the frequency-based reservoir is endowed with a hierarchy of timescales, an essential property for brain computation. This makes the selective frequency amplification an excellent mechanism for brain computation.

\begin{figure*}
\includegraphics[height=4.0in,width=6.5in]{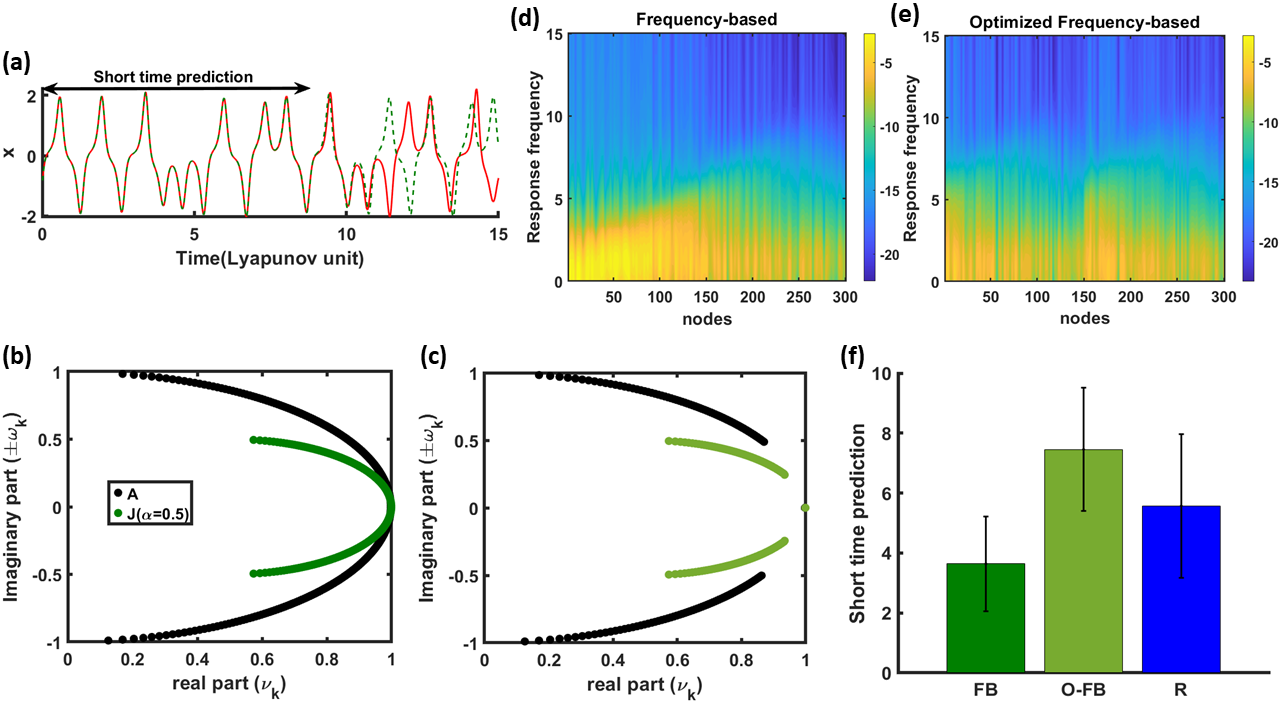}
\caption{\textbf{ Optimizing the frequency-based reservoir for short time prediction performance of the Lorenz-63.} \textbf{(a)} Short time prediction of the Lorenz evaluated in Lyapunov units (1 Lyapunov unit =0.9). \textbf{(b)} Spectrum of the default connectivity matrix $A$ (black) and the corresponding Jacobian $\text{J}$ (green)for $\alpha=0.5$ and \text{N=150}.\textbf{(c)} Spectrum of the optimized connectivity matrix $\text{A}$ (black) and the corresponding Jacobian $J$ (green) for $\alpha=0.5$. \textbf{(d)} \text{Log(power)} heat map of the frequency-based reservoir for the default connectivity in \textbf{(a)}. \textbf{(e)} \text{Log(power)} heat map of the frequency-based reservoir for the optimized connectivity in \textbf{(c)}. \textbf{(f)} Mean and standard deviation over 500 input initial conditions. The green bar represents both the default and optimized frequency-based reservoir connectivity, whereas the blue represents the random reservoir.
}
\label{FB_lorenz_optimized}
\end{figure*}

\subsection*{Optimizing the Frequency-based reservoir}
The main issue with random reservoirs is the difficulty of optimizing their performance for a specific task or target. The frequency-based reservoir selectively stores information about the input in each of its units. Choosing the intrinsic coefficients $\alpha$, $\nu_k$, and $\omega_k$ for each unit is key to achieving frequency-based reservoir performance. So far, we have chosen a default configuration in which the units have intrinsic frequencies ranging from $\omega_1=0$ to the largest value, $\omega_N$. This configuration ensures that the network is provided with all the necessary intrinsic frequencies to store all the frequency components of any given input. However, this configuration may not be optimal for specific inputs. For example, some inputs may have frequency components concentrated in a lower band, and the reservoir may not require (or require only a few) units with higher intrinsic frequencies.
On the other hand, some inputs may have a broader frequency band, requiring a reservoir with more units that have high intrinsic frequencies. Designing the reservoir so that its predictions can track the chaotic trajectory for a sufficiently long period is particularly challenging, especially with limited training data. This is known as a short-term prediction task (Fig.\ref{FB_lorenz_optimized}-\textbf{(a)}). We investigate the performance of the frequency-based reservoir on a short-time prediction task and compare it with a random reservoir. We use the chaotic Lorenz-63 system as the input. We set the reservoir size to N=301 nodes and the training set size to 1500 steps only. We found that under the default connectivity (Fig.\ref{FB_lorenz_optimized}-\textbf{(b)}), the frequency-based reservoir performs worse than the random reservoir (Fig.\ref{FB_ampl_Lorenz63}-\textbf{(f)}). The default connectivity leads to an overrepresentation of units with low intrinsic frequencies and an overprocessing of low-frequency components of the input (Fig.\ref{FB_ampl_Lorenz63}). This may explain the poor performance of the frequency-based reservoir under the default configuration. To address the overrepresentation of low intrinsic frequencies in the default configuration, we reshape the connectivity so that half of the units with low intrinsic frequencies now have the same higher intrinsic frequencies as the remaining units. This new configuration allows the reservoir to efficiently process high-frequency components of the input (Fig.\ref{FB_lorenz_optimized}-\textbf{(e)}) and significantly overperforms the random reservoir (Fig.\ref{FB_lorenz_optimized}-\textbf{(f)}). This simple optimization technique demonstrates that, unlike a random reservoir, a frequency-based reservoir can be optimized to predict a specific input.

\subsection*{The Frequency-based reservoir can forecast spatiotemporal chaotic systems}
\begin{figure*}
\includegraphics[height=3.0in,width=6.5in]{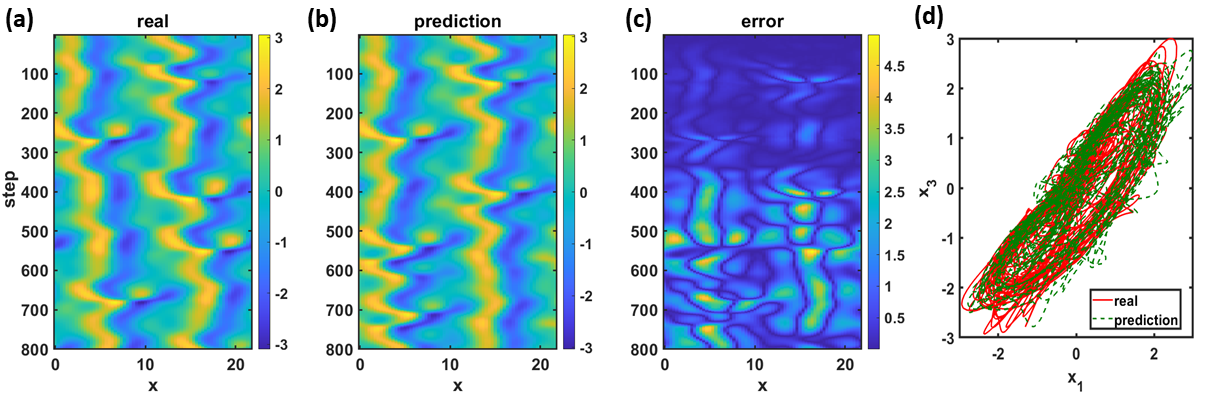}
\caption{\textbf{ Forecasting of spatiotemporal chaotic dynamical systems by a frequency-based reservoir computing.} \textbf{(a)} real data obtained after simulating the KS spatiotemporal chaotic systems and using it as input to the frequency-based reservoir. \textbf{(b)} Prediction from the frequency-based reservoir during closed-loop testing.\textbf{(c)} Error between the real data \textbf{(a)} and the prediction $\textbf{(b)}$. \textbf{(d)} Long-term prediction of the KS chaotic attractor (the third $x_3$ (y-axis) and first $x_1$ (x-axis) components of the x variable are used to build the attractor) by the frequency-based reservoir (green).
}
\label{FB_KS_forecasting}
\end{figure*}

We found that the frequency-based reservoir performs very well across a broad class of low-dimensional chaotic systems (see the supplementary information). We are now interested in how the frequency-based reservoir performs on more complex systems, such as spatiotemporal chaos. We used the Kuramoto-Shivanski as input. The results of the closed-loop forecasting are shown in Fig.\ref{FB_KS_forecasting}. The frequency-based reservoir can predict the spatiotemporal chaotic trajectory of the Kuramoto-Shivanski (Fig.\ref{FB_forecasting}-\textbf{(b)}) for a sufficiently longer epoch of time (Fig.\ref{FB_forecasting}-\textbf{(c)}) as well as its corresponding attractor (Fig.\ref{FB_forecasting}-\textbf{(d)}). Our results demonstrate the efficacy of frequency-based reservoir computing for predicting complex spatiotemporal data.

\subsection*{Properties of the frequency-based reservoir}
The performance of random reservoirs depends on the optimization of several hyperparameters, including the input amplitude, the reservoir spectral radius, and the leaky rate $\alpha$. This is usually done using several techniques, such as random search or Bayesian optimization \cite{zhai2023emergence}. Hyperparameter optimization requires substantial computational resources and relies on no practical methodology or physical principles beyond minimizing the error between the ground truth and the prediction. The process also contributes to the black box nature of random reservoirs. In a frequency-based reservoir, the hyperparameters have physical meanings and can be selected based on the theory of forced nonlinear oscillators. The leaky rate $\alpha$ shapes the eigen spectrum of the reservoir (see Fig.\ref{FB_eigenspectrum}) and therefore participates in the selective frequency amplification mechanism. When $\alpha=1$, the reservoir is an echo state network, and the eigen spectrum is directly given by the recurrent matrix $\text{A}$. The specific choice of the reservoir components $\omega_k$ and $\nu_k$ is set by the value of the spectral radius $\rho$. Specifically, the spectral radius determines the maximum frequency of the reservoir. Higher spectral radius values correspond to higher maximum frequencies, whereas lower values correspond to lower maximum frequencies (see Methods). When $\alpha$ is decreased, the eigen spectrum is shrunk, leading to an overall reduced eigen spectrum (see Fig.\ref{FB_eigenspectrum}). Therefore, the leaky rate $\alpha$ and the spectral radius both account for the reservoir's intrinsic frequencies. The network input amplitude is given by the values of the input matrix $\text{W}_{in}$, which is scaled by a coefficient $\epsilon$ (see Methods). We found that lower and intermediate values of $\epsilon$ yielded accurate predictions, whereas higher values of $\epsilon$ yielded inaccurate predictions. The inaccuracy of predictions for large values of $\epsilon$ is consistent with the destruction of the selective frequency amplification mechanism (see Fig.\ref{Single_node_ampl}). Thus, the three hyperparameters can be interpreted in the context of the theory of forced nonlinear oscillator, and their choice can be motivated by that same theory. This is in clear contrast to a random reservoir, where it's difficult to interpret the hyperparameters. A recent work \cite{zhai2023emergence} has shown that noise is an additional hyperparameter that can significantly enhance the performance of a random reservoir. We also found that the frequency-based reservoir's performance was significantly noise-dependent.
As in the random case \cite{zhai2023emergence}, we identified an optimal noise level for each task that yielded the best performance. We explain this resonant behavior by the role of noise in sustaining neural rhythms. In fact, noise is crucial for neural rhythms observed in vivo and modeled by linearly stable dynamical systems \cite{xing2012stochastic,spyropoulos2022spontaneous}. Adding noise to the frequency-based reservoir allows fast rhythms that would otherwise vanish \cite{powanwe2019determinants,powanwe2021amplitude} to contribute to input processing. In contrast, lower rhythms that would have remained coherent will be perturbed by such noise \cite{powanwe2021amplitude}. Thus, the role of noise in the frequency-based reservoir can also be interpreted through the lens of neural rhythm theory. The frequency-based reservoir provides an alternative approach in which the hyperparameter optimization step can be replaced by a transparent methodology based on oscillatory theory. 

\section{Discussion}
Many systems in physics \cite{strogatz2001nonlinear}, ecology\cite{mckane2007amplified,rozhnova2009fluctuations,mckane2005predator}, biology\cite{goldbeter1997modelling,goldbeter1997biochemical,glass2001synchronization}, and neuroscience\cite{buzsaki2006rhythms} exhibit complex oscillatory dynamics. Harnessing the oscillatory behavior of systems to perform practical computations and enable predictions has recently attracted attention. This goes from using the oscillatory behavior of Kuramoto-like dynamics for diverse computations \cite{benigno2023waves,liboni2025image} to optimizing networks of coupled harmonic oscillators for classification tasks \cite{effenberger2025functional,kramer2025brain}. In this paper, we make an additional contribution to these existing results. We have proposed a reservoir computing based on the oscillatory behavior of populations of coupled excitatory and inhibitory neurons.
In contrast to random reservoirs, in which nodes are connected at random, the frequency-based reservoir comprises excitatory-inhibitory units that are not necessarily coupled. In frequency-based reservoirs, computations occur at the unit level, and couplings among units are optional. The key insight of frequency-based reservoirs is that each unit oscillates at a specific intrinsic frequency and timescale, selectively amplifying the frequency components of the input near that frequency. By increasing the unit's intrinsic frequency from low to high, each unit will store a portion of the input information during stimulation. If the number of units and the duration of the stimulation are sufficient, a simple linear combination should be enough to recover all the information contained in the input and therefore reconstruct it accurately.\\

The frequency-based reservoir introduced here offers several advantages in terms of biological realism. First, neural rhythms recorded across multiple brain regions and frequency bands are believed to be involved in many cognitive processes, including communication via coherence \cite{fries2005mechanism,fries2015rhythms}, attention \cite{bosman2012attentional}, memory, and learning \cite{wang2010neurophysiological}. Second, it has been shown that the brain possesses a hierarchy of time scales \cite{runyan2017distinct,siegle2021survey,murray2014hierarchy,spitmaan2020multiple,manea2022intrinsic,hasson2008hierarchy}, in which early sensory regions process input rapidly. In contrast, other regions interact with inputs on much slower timescales. Like the brain, our frequency-based reservoir exhibits intrinsic oscillatory dynamics across different frequency bands and processes inputs according to a hierarchy of timescales.
Additionally, the Wilson-Cowan model is a well-established model for brain dynamics, used to simulate oscillatory neural recordings across various species and frequency bands. The selective frequency amplification mechanism identified here may reflect a biologically plausible way the brain processes and stores information to support accurate future predictions. It is biologically plausible that the coordinated activity of neurons (here reflected in the oscillatory dynamics of each Wilson-Cowan unit) in each brain region plays an essential role in processing and storing information about sensory inputs. Moreover, the frequency-based reservoir provides a transparent interpretation of what is happening inside the reservoir, a property that is invaluable for efficient optimization.\\

Random reservoirs have demonstrated strong performance in predicting chaotic time series \cite{tanaka2019recent} with a minimal number of learnable parameters. However, they suffer from the curse of interpretability since the reservoir is seen as a black box. Previous works have sought to shed light on the black-box nature of random reservoirs by designing more interpretable and biologically realistic reservoirs \cite{srinivasan2025boosting} or by connecting random reservoirs to nonlinear autoregressive processes \cite{gauthier2021next}. The frequency-based reservoir leverages the intrinsic oscillatory behavior and the hierarchy of timescales observed across brain areas. Our approach is grounded in the theory of nonlinear oscillators driven by complex, multifrequency periodic inputs. We showed that the dynamics of the frequency-based reservoir could be interpreted as those of a single forced nonlinear oscillator \cite {lindner2009local}. By varying the oscillator intrinsic frequency and timescale, we can infer which input frequency components are processed and stored at a given node/unit of the reservoir. This provides interpretability for the frequency-based reservoir, which, to our knowledge, has not been observed in prior works. The frequency-based reservoir has a small number of connections. For a reservoir with $\text{N}$ units, the frequency-based reservoir has $\text{2N-1}$ nodes and $\text{4N-1}$ connections. It is equivalent to an ensemble of small, independent reservoirs (each with two nodes) that process a portion of the same input. The absence of couplings among the units in the reservoir makes it readily amenable to experimental realization and to applications in neuromorphic computing. This enables the design and fabrication of reservoirs with oscillatory units based on the selective frequency-amplification mechanism described here.\\

 A recent line of works has used the harmonic oscillator as a basic unit for computations. Results from networks of coupled harmonic oscillators are promising \cite{rusch2020coupled,rusch2022graph} and indicate that they can outperform neural networks that lack intrinsic oscillatory dynamics. Harmonic oscillatory recurrent networks (HORN) have been shown to outperform non-oscillatory networks in pattern recognition tasks \cite{effenberger2025functional}. The HORN is based on a large number of randomly selected parameters and uses a random connectivity matrix. It was trained using backpropagation through time.
In contrast to our frequency-based reservoir, the complexity of HORN and its training procedure may obscure the exact computational mechanisms at play. A reservoir computing approach based on networks of coupled random oscillators has been proposed in \cite{ceni2024random}. The architecture is similar to the HORN architecture with a random connectivity matrix. As in our case, only the readout matrix is trained. However, their reservoir is significantly different from our frequency-based reservoir. Our reservoir is fully interpretable, does not rely on the random choice of intrinsic parameters or a random connectivity matrix. In contrast with the reservoir in \cite{ceni2024random}, the dynamics of each unit can be fully understood independently of those of other units. We have proposed a mechanism that our frequency-based reservoir uses for computation and prediction. Another recent work \cite{kramer2025brain} has used reservoir computing based on the damped harmonic oscillator for brain-rhythm classification and handwritten and spoken digit classification. They proposed a resonant mechanism that their reservoir uses for classification. Although their resonant behavior shares some similarities with selective frequency amplification, our mechanism is more general. The selective frequency amplification operates by amplifying a spectrum of continuous frequencies near the oscillator's intrinsic frequency. It extends beyond periodic inputs and generalizes to high-dimensional chaotic dynamics. Moreover, our frequency-based reservoir can be designed and optimized to predict a specific input. A property that has not been shown in previous oscillatory-based neural networks.\\

The dynamics of the damped harmonic oscillator and the Wilson-Cowan model can overlap in the parameter regime in which the latter has complex-conjugate eigenvalues with negative real parts. However, the two models are different. A single Wilson-Cowan neuron is a nonlinear dynamical system, and its response to inputs can be highly complex. A single damped harmonic oscillator is a linear dynamical system. The nonlinearity is introduced through the coupling with other units \cite{ceni2024random,effenberger2025functional}. While a single Wilson-Cowan unit is already nonlinear, in most damped-harmonic-oscillator neural networks, the nonlinearity is emergent. Thus, computations in our frequency-based recurrent network happen at the unit level, and coupling between the units is optional. This opens interesting questions about computations with the frequency-based reservoir. For example, what is the minimum number of units required to predict a given input? Which input frequency components should be accurately processed by the frequency-based reservoir to increase the performance in short/long-term prediction? The answer to the first question can be anticipated for periodic inputs as $\text{2N}\leq \text{2p}$, where $\text{N}$ is the number of units and $\text{p}$ the number of  harmonics of the input.  For chaotic inputs, the question remains open. For the second question, the optimization procedure (Fig.\ref{FB_lorenz_optimized}) suggests that the frequency-based reservoir should accurately process the fast components of the chaotic inputs to improve short-term prediction.\\

Another distinction between a frequency-based reservoir and a neural network based on a damped harmonic oscillator concerns the role of noise. While noise was optional \cite{kramer2025brain} or seen as a robustness measure\cite{effenberger2025functional}, in coupled damped harmonic oscillator networks, we found that noise was critical for accurate short and long-term prediction in the frequency-based neural network. The effect of noise was observed for chaotic inputs with fast (e.g., the Mackey-Glass) and low-frequency components. Previous work has identified a noise-induced resonance mechanism in random reservoirs \cite{zhai2023emergence}. Here, we argue that noise is critical because it induces oscillations in units with a fast intrinsic frequency and perturbs rhythms in units with a low inherent frequency \cite{powanwe2021amplitude}. In fact, units with large intrinsic frequencies decay quickly, and noise acts as a random force that sustains them \cite{burns2011gamma,xing2012stochastic,powanwe2019determinants}. The noise-induced rhythms are filtered noise signals suitable for predicting the fast components of the inputs.
On the other hand, units with small intrinsic frequencies remain coherent for a very long epoch of time, similar to self-sustained or limit-cycle oscillations. Such coherent, self-sustaining rhythms could be less flexible in efficiently processing and storing external inputs. Noise may act as a force that breaks the self-sustained or coherent nature of such low-frequency oscillations \cite{powanwe2021amplitude}, enabling them to process and store information about external inputs efficiently.

\section{Methods}
\subsection*{Frequency-based recurrent layer}
The construction of the recurrent matrix $\textbf{A}$ in Equation \ref{reservoir_dynamics} was motivated by the oscillatory dynamics of a network of connected Wilson-Cowan units (see Supplementary Information). The dynamics of a single Wilson-Cowan unit have been extensively studied \cite{wilson1972excitatory,wallace2011emergent}. In the regime in which its complex-conjugate eigenvalues have negative real parts, the dynamics decay to zero. However, adding noise sustains the rhythms \cite{powanwe2019determinants,powanwe2021brain,powanwe2021amplitude}. This is especially relevant for fast rhythms. We found that noise was essential for predicting chaotic time series. For that reason, the reservoir's input was corrupted by noise. This leads to the following effective dynamics for the frequency-based reservoir.

\begin{equation}
\label{reservoir_noisy_dynamics}
\text{r(t+1)}=(1-\alpha)\text{r(t)}+\alpha \tanh\bigg(\text{Ar(t)}+\text{W}_\text{in}\bigg(\text{u(t)}+\sigma \xi(t)\bigg)\bigg),
\end{equation}

where $\xi(t)$ is a time-dependent Gaussian white noise with zero mean and unit variance, the coefficient  $\sigma$  controls its intensity.
 The matrix  $\textbf{A}$ was initialized by first choosing the spectral radius $\rho$, and then selecting $\text{N}$ angular frequencies (or imaginary parts) $\omega_k$ equally spaced from a minimum value $0\leq\omega_{min}$ to a maximum value $\omega_{max}\leq \rho$. The eigenvalue real parts were then deduced as $\nu_k=\sqrt{\rho^2-\omega^2_k}$. Furthermore, we arranged the units so that the first unit has the lowest angular frequency, $\omega_{min}$, and the last unit, $\text{N}$, has the largest angular frequency, $\omega_{max}$.
 With this choice, each unit is a small reservoir of two connected nodes and a spectral radius of $\rho$.  The frequency-based reservoir is an ensemble of $\text{N}$ small reservoirs (each having two nodes) with the same spectral radius, $\rho$. Furthermore, the frequency-based reservoir is endowed with a hierarchy of timescales (units with large indices $k$ have larger angular frequencies $\omega_\text{k}$ and appear earlier in the hierarchy). In contrast, those with small $k$ have smaller angular frequencies $\omega_\text{k}$ and appear later ), a key property of brain computation \cite{runyan2017distinct,siegle2021survey,murray2014hierarchy,spitmaan2020multiple,manea2022intrinsic,hasson2008hierarchy}.
The Jacobian of the Equation. \ref{reservoir_noisy_dynamics} is obtained by linearization, leading to
\begin{equation}
\label{Jacobian_equation}
    \text{J}=(1-\alpha)\text{I}+\alpha \text{A},
\end{equation}
where $\textbf{I}$ is an $\text{2Nx2N}$ identity matrix. The corresponding eigenvalues are complex conjugate pairs given as $(1+\alpha)+\alpha \nu_k \pm \text{j}\alpha \omega_k$, with $\text{k=1,...,N}$ and $\text{j}=\sqrt{-1}$ the pure imaginary number.

The oscillatory frequency of each unit is obtained as 

\begin{equation}
    \label{Intrinsic_frequency}
    \text{f}_\text{k}=\dfrac{\alpha\omega_k}{2\pi}.
\end{equation}
The coefficient $\alpha$ scales the value of the intrinsic frequency. When $\alpha=1$, we have an echo state and $\mathbf{A}$ and $\mathbf{J}$ coincide. In contrast to a random reservoir, where the role of $\alpha$ is speculative and often associated with reservoir memory, here we can clearly identify $\alpha$ as a factor that scales the network's intrinsic frequencies, thereby contributing to the selective frequency-amplification mechanism.  The matrix $\text{W}_\text{in}$ represents a feedforward layer that maps the noise-corrupted input $\text{u(t)}+\sigma \xi(t)$ to the recurrent layer. Its size is  $\text{(d, 2N-1)}$ where d is the dimension of the input and \text{2N-1} that of the reservoir. Precisely, it was chosen as $\text{W}_\text{in}=\epsilon \text{(2*U(d, 2N-1)-1 )}$, where \text{U} is a \text{(d, 2N-1)} matrix with components generated from a uniform distribution between 0 and 1. This makes the values of $\text{W}_\text{in}$ random numbers uniformly generated between $-\epsilon$ and $\epsilon$. $\text{W}_\text{in}$ is the only matrix that was randomly generated. With this formulation, the effective noise intensity is $\epsilon\sigma$ scaled by the corresponding random value generated by the matrix $
\text{2*(U(d,2N-1)-1 )}$. The output matrix $\text{W}_\text{out}$ was obtained by minimizing the quantity $  \| \mathbf{W_{out}r(t)-y(t)} \|_2$. Previous work in reservoir computing has employed Tikhonov regularization to mitigate overfitting. Here, we impose a different constraint. In addition to minimizing the quantity $  \| \mathbf{W_{out}r(t)-y(t)} \|_2$ we consider the solution $\text{W}_\text{out}$ that also minimizes the norm  $ \| \mathbf{W_{out}}\|_2 $. In Matlab 2024a, we used the command \text{lsqminnorm} to obtain $W_{out}$. During testing, we introduce a feedback loop by replacing the input $\text{u(t)}$ with $\text{W}_\text{out}\text{r(t)}$. The dynamics of the reservoir during closed-loop testing are a self-evolving discrete dynamical system given by

\begin{equation}
\label{reservoir_dynamics_testing}
\text{r(t+1)}=(1-\alpha)\text{r(t)}+\alpha \tanh\bigg(\text{Ar(t)}+\text{W}_\text{in}\text{W}_\text{out}\text{r(t)}\bigg).
\end{equation}

During training and closed-loop testing, the even nodes of the reservoir were further squared. This is a common practice when the reservoir uses the $\tanh$ nonlinearity. We applied this approach to all inputs in the main text.

\subsection*{Forecasting the Mackey-Glass chaotic system}
The input in Fig.\ref{FB_forecasting}-\textbf{(a-c)} was generated by the Mackey-Glass dynamical system \cite{mackey1977oscillation} whose dynamic is given as

\begin{equation}
    \dfrac{du}{dt}=\dfrac{\beta u(t-\tau)}{1+u^k(t-\tau)}-\gamma u(t),
\end{equation}
with the following coefficients $\beta=0.2$,$\gamma=0.1$, $k=10$ and $\tau=17$. We used a recurrent layer with $N=301$ units, which corresponds to \text{2N-1=601} units. The input was simulated using a fourth-order Runge-Kutta scheme with a step size of \text{dt = 0.01}. A total of $T=5.10^3+5.10^3+10^5$ samples were generated. The first $ 5.10^3$ were discarded as transient, the next $5.10^3$ were used as input during the reservoir washup period, and the remaining $10^5$ were used for training. The self-evolving reservoir during the closed-loop testing period was simulated for $T=2.10^5$ steps to construct the attractor in Fig.\ref{FB_KS_forecasting}-\textbf{(c)}. Further, the noise intensity was chosen as $\sigma=10^{-3.5}$, $\alpha$ was set to $0.5$ and the input strength $\epsilon=0.05$. The same input configuration was used for the random reservoir. The random reservoir was generated as a random matrix with symmetrically distributed coefficients and degree $\text{q}$. The value of q was chosen as \text{q = 0.00525} to ensure that the random and frequency-based reservoirs have approximately the same number of connections.

\subsection*{Forecasting the Lorenz-63 chaotic system}
The Lorenz-63 chaotic input was generated using the following dynamical system \cite{lorenz2017deterministic}

\begin{align}
    \dfrac{dx}{dt}&=10(y(t)-x(t))\\
    \dfrac{dy}{dt}&=x(t)(28-z(t))-y(t)\\
    \dfrac{dz}{dt}&=x(t)y(t)-\dfrac{8}{3}z(t)
\end{align}
and the corresponding input was constructed as $u=[x(t),y(t),x(t)]$. In Fig.\ref{FB_KS_forecasting}-\textbf{(d-f)} we used the Lorenz-63 as input. The recurrent layer size was the same as in the Mackey-Glass case. The Lorenz-63 was simulated with a Fourth-order Runge-Kutta algorithm. All the other training parameters were almost the same as in the Mackey-Glass case, apart from the noise strength $\sigma=10^{-7}$ and the training length $T=2.10^5$. For Fig.\ref{FB_lorenz_optimized}, we significantly reduced the training size to $T=1500$ steps, the network size was also reduced to $N=301$, and the noise intensity was modified to $\sigma=10^{-6}$. The same parameters were chosen for the random recurrent layer, with the degree $q=0.0067$.

\subsection*{Forecasting the spatiotemporal Kuramoto-Sivashinki chaotic system}
In Fig.\ref{FB_KS_forecasting}, the input we chose was the Kuramoto-Sivashinsky. A one-dimensional model of instabilities in flames is described as
\begin{equation}
\label{Kuramoto_Sivashinki_dynamics}
    \dfrac{\partial u}{\partial t}=-\dfrac{\partial^2u}{\partial x^2}-\dfrac{\partial^4 u}{\partial x^4}-u\dfrac{\partial u}{\partial x}
\end{equation}
The variable $u=u(x,t)$ describes a spatiotemporal behavior. We simulate Equation.\ref{Kuramoto_Sivashinki_dynamics} on the spatial dimension x from $0$ to $L=22$. We chose sinusoidal initial condition $u(x,0)=\sin\bigg(2\pi\dfrac{x}{L}\bigg)$ and periodic boundary condition. The temporal and spatial resolutions were set to $dt=0.25$ and $dx=\dfrac{1}{3}$, respectively. The reservoir size was set to $N=2201$ and initialized with the default configuration given in Fig.\ref{FB_eigenspectrum}. The input was corrupted by Gaussian white noise with strength $\sigma=10^{-1}$. The other parameters were the same as in the Mackey-Glass and Lorenz-63 cases. The training phase consists of a total of $\bm{T}=2.10^5$ steps. In the closed loop testing, we self-evolve the reservoir for $T=10^4$ steps to generate the attractor shown in Fig.\ref{FB_KS_forecasting}-\textbf{(d)}.

\subsection*{Power spectra}

All the power spectra in the main text and the supplementary information were generated using the $pwelch$ command in Matlab 2024a. The heat maps were generated by taking the logarithm of the power spectra for all nodes in the recurrent layer.

\subsection*{ The Wilson-Cowan frequency-based reservoir}
We first developed our analysis and obtained our results by using the Wilson-Cowan model. The discrete dynamics of the Wilson-Cowan recurrent layer are given by

\begin{align}
    \label{Discrete_Exc_1}
    E_{k}(t+1)&=(1-\Delta t\alpha_{e})E_{k}(t)+\Delta t\beta_{e}\big(1-E_{k}(t)\big)f(s_{E_{k}})\\
    \label{Discrete_Inhi_1}
    I_{k}(t+1)&=(1-\Delta t\alpha_{i_k})I_{k}(t)+\Delta t\beta_{i}\big(1-I_{k}(t)\big)f(s_{I_{k}})
\end{align}
The total synaptic inputs to the E and I populations are given by

\begin{align}
    \label{Discrete_Input_E}
    s_{E_{k}}(t)&=W_{ee}^{k}E_{k}-W_{ei}^{k}I_{k}+h_{e}^{k}+\sum_{l=1}^{N}W_{kl}E_{l}(t)
    +\sum_{p=1}^{P}W_{in}^{kp} u_{p}(t)+\sigma_{e_k}\xi_{e_{k}}(t)\\
     \label{Discrete_Input_I}
    s_{I_{k}}(t)&=W_{ie}^{k}E_{k}-W_{ii}^{k}I_{k}+h_{i}^{k}+\sum_{l=1}^{N}W_{kl}E_{l}(t)
    +\sum_{p=1}^{P}W_{in}^{kp} u_{p}(t)+\sigma_{i_k}\xi_{i_{k}}(t)\,.
\end{align}
Each unit $(E_k, I_k)$ has its own oscillatory dynamics   described by the coefficients $W^k_{ee}$ $W^k_{ii}W^k_{ei}$,$W^k_{ie}$ $\alpha_e$,$\alpha_{ik}$, $\beta_e$,$\beta_i$, ,$h^k_e$,$h^k_i$,$\sigma_{e_k}$ and $\sigma_{i_k}$. The biological meanings of these parameters are provided in the supplementary material. The parameter $\Delta t$ is a step-size that plays the same role as $\alpha$. The nonlinear response function $f$ is the sigmoid function $f(x)=\dfrac{1}{1+e^{-x}}$. The 13 parameters make the single-unit Wilson-Cowan model difficult to modify and optimize for those unfamiliar with it. However, we found that the parameters $\alpha_{i_k}$ and $W^k_{ee}$ were useful for modifying the intrinsic frequency of each unit (see Supplementary Information). The units are coupled through the matrix $\text{W}$ and receive the input $\text{u(t)}$ through the feedforward input layer $\text{W}_\text{in}$. The coupling targets both nodes of each unit, but only the E node projects to other units. In the supplementary information, we performed a detailed analysis of the Wilson-Cowan model described here. We chose two coupling matrices, a ring and a distance-dependent one. For both matrices, the main effect was to shape the unit intrinsic frequencies. This led us to conclude that computation occurs at the unit level and that coupling is an additional parameter that contributes to prediction by shaping each unit's intrinsic dynamics. In fact, we found that even a network of uncoupled Wilson-Cowan units (\text{W=0}) could make accurate predictions. However, each Wilson-Cowan unit has 13 coefficients, making it difficult to optimize for predictive performance. However, the recurrent layer described in the main text is easier to control and optimize than the Wilson-Cowan model.
To obtain the output matrix $\text{W}_\text{out}$, we collect all the reservoir states as 

\begin{equation}
\label{reservoir_state_WC}
   r(t)= \begin{bmatrix}
  E_{2k-1} \\
  I_{2k} 
  \end{bmatrix}  \,, k=1,2,3,...N,
\end{equation}

and minimize the quantity Equation.\ref{loss_function}. In contrast to the $\tanh$ nonlinearity, the sigmoid nonlinearity does not require squaring the even reservoir state components. During closed-loop testing, the input for the next step is replaced by the prediction from the current step. The Wilson-Cowan reservoir, as described here, implements the selective amplification mechanism to store information about the inputs and uses that information for prediction, as described in the main text (see Supplementary information).

\nocite{*}

\bibliography{Biblio}



\end{document}